\documentclass{article}
\usepackage{epsfig}
\usepackage{alltt}
\usepackage{amssymb}

\def\w#1{{\it #1}} 

\def\i{{\mbox{\(^{-1}\)}}}

\def\la{{\mbox{\(\lambda\!\)}}}

\def\union{\cup}

\def\set#1{{\{#1\}}}
\def\myem{\sc}

\def\freegroup#1{{F(#1)}}
\def\nm#1{{\mbox{NM}(#1)}}
\def\ng#1{{\mbox{NG}(#1)}}

\def\conceptdef#1{{\sc #1}}

\def\gcs{\mbox{GCS}}
\def\gcsa{\mbox{GCSA}}

\def\atoms{{\mbox{\(V \union V\i\)}}}

\newtheorem{theorem}{Theorem}

\def\fg{{\mbox{$\freegroup{V}$}}}

\def\sub#1{{\mbox{\(_{#1}\)}}} 

\def\rra{{\mbox{\(\:\mapsto\:\)}}} 

\def\ra{{\mbox{\(\:\rightarrow\:\)}}} 

\def\squareforqed{\hbox{\rlap{$\sqcap$}$\sqcup$}}
\def\qed{\ifmmode\squareforqed\else{\unskip\nobreak\hfil
\penalty50\hskip1em\null\nobreak\hfil\squareforqed
\parfillskip=0pt\finalhyphendemerits=0\endgraf}\fi}

\def\x{{\it x}}
\def\y{{\it y}}
\def\z{{\it z}}

\def\multiset#1{\set{#1}}
\def\multirelator#1{{\langle#1\rangle}}
\def\multiconjugates#1{{\mbox{MC}}{(#1)}}

\def\myfigref#1{{Fig.~\ref{#1}}}

\def\fg{{\mbox{$\freegroup{V}$}}}

\def\ggone{{\tt G-Grammar}}
\def\ggtwo{{\tt G-Grammar'}}

\def\mr{{\mit mr}}

\def\MC{{\mbox{MC}}}

\def\ss{{\mit ss}}
\def\ps{{\mit ps}}

\def\Z{{\mathbb{Z}}}  

\begin{document}

\title{Some remarks on the geometry of grammar}

\author{
Marc Dymetman\\[1ex]
Xerox Research Centre Europe\\
6 chemin de Maupertuis\\
38240 Meylan, France\\[1ex]
{\tt dymetman@xrce.xerox.com}}

\date{{February 1999}}

\maketitle

\begin{abstract}
  This paper, following \cite{Dymetman:1998c}, presents an approach to
  grammar description and processing based on the geometry of {\em
    cancellation diagrams}, a concept which plays a central role in
  combinatorial group theory \cite{LyndonSchuppe:1977}. The focus here
  is on the geometric intuitions and on relating group-theoretical
  diagrams to the traditional charts associated with context-free
  grammars and type-0 rewriting systems. The paper is structured as
  follows. We begin in Section 1 by analyzing charts in terms of
  constructs called {\em cells}, which are a geometrical counterpart
  to rules. Then we move in Section 2 to a presentation of
  cancellation diagrams and show how they can be used computationally.
  In Section 3 we give a formal algebraic presentation of the concept
  of {\em group computation structure}, which is based on the standard
  notions of free group and conjugacy. We then relate in Section 4 the
  geometric and the algebraic views of computation by using the
  fundamental theorem of combinatorial group theory
  \cite{Rotman:1994}. In Section 5 we study in more detail the
  relationship between the two views on the basis of a simple grammar
  stated as a group computation structure.  In section 6 we extend
  this grammar to handle non-local constructs such as relative
  pronouns and quantifiers. We conclude in section 7 with some brief
  notes on the differences between normal submonoids and normal
  subgroups, group computation versus rewriting systems, and the use
  of group morphisms to study the computational complexity of parsing
  and generation.
\end{abstract}

\bibliographystyle{plain}

\section{Introduction: grammar and geometry} \label{Grammar and geometry}

\begin{figure}[ht]
  \begin{center}
    \epsfig{file=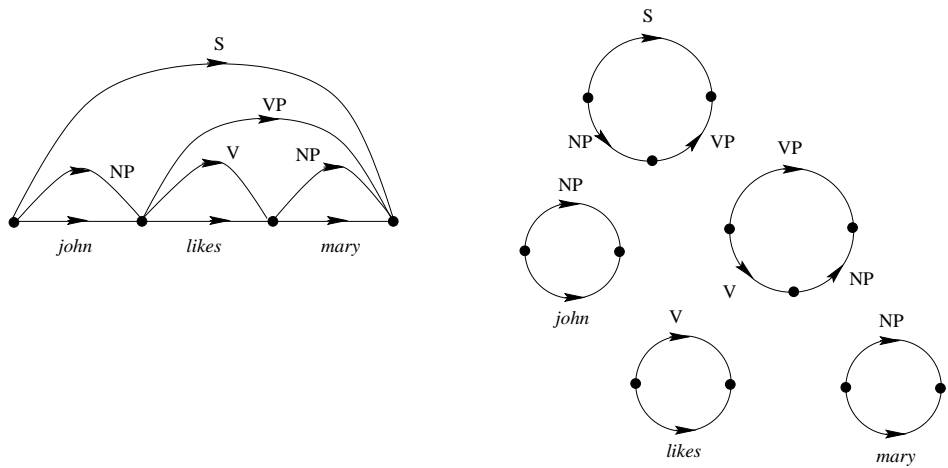,scale=0.8}
    \caption{A context-free chart and its decomposition into cells.}
    \label{fig:chart}
  \end{center}
\end{figure}

In the drawing on the left of Figure \ref{fig:chart}, we give an
example of a simple context-free chart, where we have assumed the
edges to be conventionally oriented from left to right.

Seen in geometric terms, this chart is an oriented planar graph
dividing the plane in five internal regions and one external one. Each 
of the internal regions can be transformed into a disk by a
topological transformation of the plane preserving oritentation. The
five corresponding circular graphs, or \conceptdef{cell}s, are shown on 
the right. Relative to one of these cells, an edge can be considered
oriented positively if it goes clockwise, or negatively if it goes
counter-clockwise. Thus, the NP edge of the S-NP-VP cell is oriented
negatively, while the NP edge of the NP-john cell is oriented
positively.

It is then an obvious observation that the chart on the left can be
obtained by a process of ``stitching'' together the five cells along
{\em identically labelled but inversely oriented} edges. In the course
of this operation, each cell undergoes an orientation-preserving
topological deformation.

The cells of Figure \ref{fig:chart} provide a geometric presentation
of context-free productions. Each cell has the property that it has
exactly one positively oriented edge, labelled with a nonterminal, and
one or several negatively oriented edges, labelled with nonterminals or
terminals. If, in the context-free chart, one chooses a point in each
internal region and draws lines between two points belonging to
adjacent regions,%
\footnote{This construct is related to the notion of {\em dual} graph
  of a planar graph \cite{Berge:85}.}  then one obtains a planar graph
which is tree-like, and which corresponds to the usual notion of
derivation tree in a context-free grammar.

\begin{figure}[ht]
  \begin{center}
    \epsfig{file=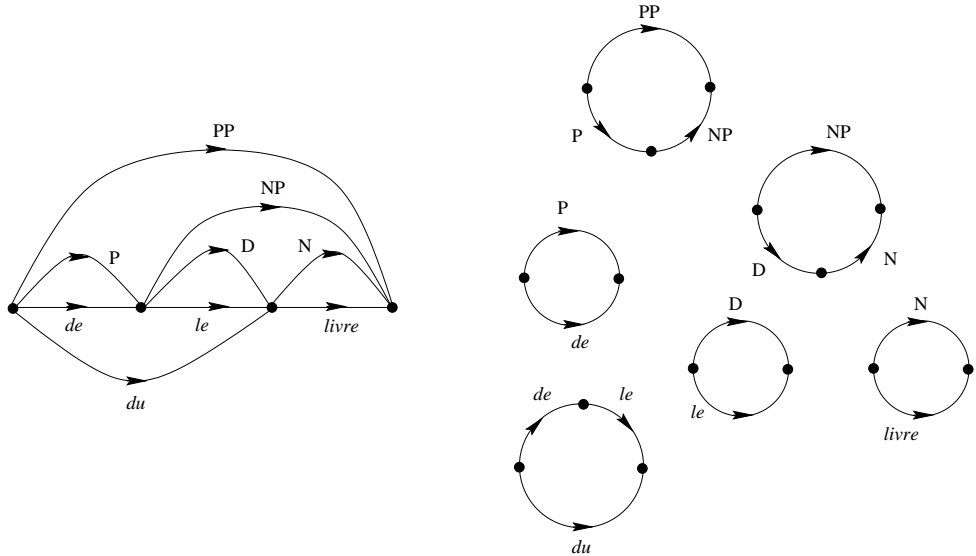,scale=0.8}
    \caption{A ``type-0 chart'', or ``Q-system chart'', and the
      corresponding cells.}
    \label{fig:qsys}
  \end{center}
\end{figure}

If, instead of a context-free grammar, one considers a general type-0
grammar, charts can be generalized to graphs which do not have this
``tree-like'' property. An illustration of such a case is given in
Figure \ref{fig:qsys}, which shows a derivation of the French
prepositional phrase ``du livre'', involving the contraction ``du'' of
the preposition ``de'' with the determiner ``le''. The graph on the
left could be called a ``type-0 chart'', or a ``Q-system
chart'', in reference to \cite{Co71}.%
\footnote{Q-systems were a direct predecessor of Prolog. Because of
  the symmetry between input and output, the formalism could be
  applied to both parsing and generation in machine translation and
  was a pioneer of grammar reversibility.}

The cells associated with this chart are shown on the right. In
contrast to the context-free case, one of the cells now contains two
positively oriented edges.

In the computation systems we are proposing here, we will consider
cells with an arbitrary number of positively or negatively oriented
edges to be the building blocks from which computations are built. Our 
formalization will rely on the mathematical concept of a
\conceptdef{cancellation diagram}.

\section{Cancellation Diagrams}

\subsection{Diagrams}

Cancellation diagrams were introduced in 1933 by Van Kampen
\cite{VanKampen:1933} and have been playing since the seventies an
increasing role in combinatorial group theory
\cite{LyndonSchuppe:1977,Rotman:1994,DJohnson:1997}.

{\bf Definition}. A \conceptdef{cancellation diagram}, or simply
\conceptdef{diagram}, over the vocabulary $V$ is a finite graph which
is: (1) planar, that is, embedded in the plane in such a way that two
edges can only intersect at a vertex; (2) connected; (3) directed,
that is, the edges carry an orientation; (4) labelled, that is, each
edge carries a label taken in $V$. In the limit, a graph consisting of
a single vertex is also considered to be a diagram.

A diagram separates the plane in $n+1$ connected open sets: the
exterior (set of points that can be connected to a point at infinity
without crossing an edge), and $n$ open internal regions, called
\conceptdef{cell region}s, each consisting of points which can be
connected without crossing an edge, but which are separated from the
exterior. The set of points in the plane which belong to an edge or
to an internal region of the diagram is called the \conceptdef{locus}
of the diagram. The locus of a diagram is always a connected and
simply connected (no holes) closed region of the plane.  An edge whose
points are not in the closure of a cell region is called a
\conceptdef{thin} edge of the diagram.

\begin{figure}[ht]
  \begin{center}
    \epsfig{file=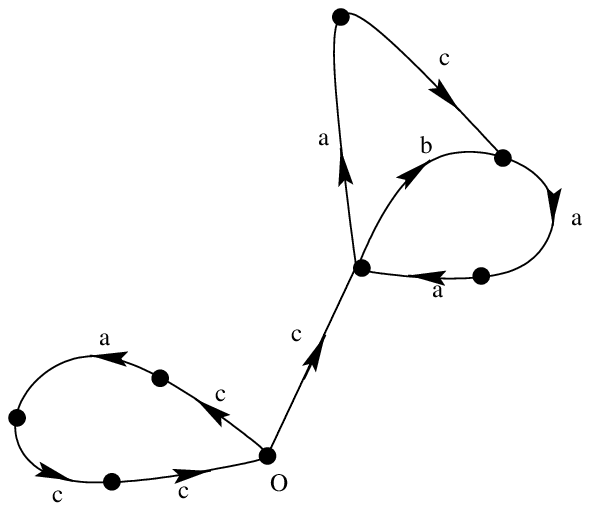,scale=0.8}
    \caption{A diagram. A boundary of this diagram is $c\i c\i a\i c\i 
      c a c a a c\i$.}
    \label{fig:first-diagram}
  \end{center}
\end{figure}

An example of a diagram over the vocabulary $\set{a,b,c}$ is given in
\myfigref{fig:first-diagram}. This diagram is made up of three cells
and contains one thin edge.

The \conceptdef{boundary} of a cell is the set of edges which constitute its
topological boundary. The boundary of a diagram is the set of edges
which are such that all their points are connected to the exterior.  

If one choses an arbitrary vertex (such as O in the figure) on the
boundary of a diagram, and if one moves on the boundary in a
conventional clockwise fashion, then one collects a list of edges
which are either directed in the same way as the movement, or contrary
to it. By producing a sequence of labels with exponent $+1$ in the
first case, $-1$ in the second case, one can then construct a word over the
vocabulary \atoms (that is, a word in the {\em free group} over $V$,
see below); this word is said to be a \conceptdef{boundary word} of
the diagram.

\subsection{Reduced diagrams}

\begin{figure}[ht]
  \begin{center}
    \epsfig{file=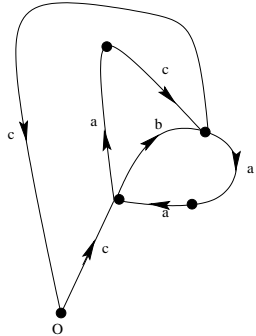,scale=0.8}
    \caption{A reduced diagram. The diagram of the previous figure was
      not reduced because of the two $c$ edges outgoing from vertex $O$.}
    \label{fig:reduced-diagram}
  \end{center}
\end{figure}

We will say that a diagram is \conceptdef{reduced} if there does not
exist a pair of edges with a common vertex $O$, with the same label,
oriented oppositely relative to $O$ (that is, both edges point towards
$O$ or both point from $O$), and such that at least one of the two
``angles'' formed by the two edges is ``free'', that is, does not
``contain'' another diagram edge (see \myfigref{fig:reduced-diagram}).

\subsection{Diagrams as computational devices} \label{Diagrams as computational devices}

Suppose that one is given a fixed set of cells $C$ over a vocabulary
$V$. Let's consider the following generative process for producing
diagrams (only informally described here):
\begin{enumerate}
\item Initialize the diagram by choosing a point $O$ in the plane;
\item Iterate an arbitrary number of times the following procedure any 
  of the following steps:
  \begin{itemize}
  \item Add a new oriented labelled edge to the exterior of the
    current diagram by connecting it to one vertex on the boundary of
    the diagram;
  \item Add a new cell taken from $C$ to the exterior of the current
    diagram, either by connecting one vertex of the cell to one vertex on
    the boundary of the diagram or by ``pasting'' the cell to the
    boundary of the diagram along consecutive edges having the same
    labels and arrow directions;
  \item Reduce the current diagram by ``folding together'' two
    adjacent edges on its boundary which have the same label but
    opposite directions (informally, the deformation involves
    gradually reducing the external ``angle'' between the two edges
    until they are identified, that is, the folding is towards the
    interior of the diagram);
  \end{itemize}
\end{enumerate}

It is easy to prove that this process generates exactly the set $D$ of
diagrams $d$ over $V$ such that all the cells of $d$ are elements of
$C$. The boundary words of elements of $D$ can be seen as {\em coding}
results of the computations determined by the ``specification'' $C$.

For instance, going back to the example of Figure \ref{fig:chart}, the
boundary word $S \: \w{mary}\i \w{likes}\i \w{john}\i$ is the result of a
computation over the specification $C$ consisting of the five circular
cells in the figure. Any word thus obtained which is of the form $S \:
{\w{word}_n}\i \ldots {\w{word}_1}\i$ can be seen as coding the fact
that ${\w{word}_{1}} \ldots {\w{word}_n}$ is a sentence relative to
the ``grammar'' specified by $C$.

This informal notion of computation with diagrams will now be made
more precise by turning to its algebraic counterpart, {\em group
  computation structures}.

\section{Computation in groups}

We start by quickly reviewing some basic concepts of group theory
before turning to group computation structures.

\subsection{Groups, monoids, normal subsets} \label{Groups, monoids, normal subsets}

A \conceptdef{monoid} $M$ is a set $M$ together with a product $M
\times M \ra M$, written $(a, b) \rra ab$, such that:

\begin{itemize}
\item This product is associative;
\item There is an element $1 \in M$ (the neutral element) with $1a =
  a1 = a$ for all $a \in M$.
\end{itemize}

A \conceptdef{group} is a monoid in which every element $a$ has an
inverse $a\i$ such that $a\i a = a a\i = 1$.

A \conceptdef{submonoid} of $G$ is a subset of $G$ containing $1$ and
closed under the product of $G$. A \conceptdef{subgroup} of $G$ is a
submonoid of $G$ which is closed under inversion.

Two elements $x,x'$ in a group $G$ are said to be
\conceptdef{conjugate} if there exists $y\in G$ such that $x' =  y x y\i$.

A subset (resp. a subgroup, a submonoid) of a group $G$ is said to be
a \conceptdef{normal subset} (resp. \conceptdef{normal subgroup},
\conceptdef{normal submonoid} of $G$ iff when it contains $x$, it
contains all the conjugates of $x$ in $G$.

If $S$ is a subset of $G$, the intersection of all normal submonoids
of $G$ containing $S$ (resp.\ of all subgroups of $G$ containing $S$)
is a normal submonoid of $G$ (resp.\ a normal subgroup of $G$) and is
called the \conceptdef{normal submonoid closure} $\nm{S}$ of $S$ in
$G$ (resp.\ the \conceptdef{normal subgroup closure} $\ng{S}$ of $S$
in $G$).

The notion of normal subgroup is central in algebra. For our purposes
here, the less usual notion of normal submonoid will be the more
important notion.

\subsection{The free group over V.} 

Let's consider an arbitrary set $V$, called the
\conceptdef{vocabulary}, and let's form the so-called {\myem set of atoms
  on $V$}, which is notated $V \union V\i$ and is obtained by taking
elements $v$ in $V$ as well as the formal inverses $v\i$ of these
elements.

We now consider the set $\freegroup{V}$ consisting of the empty
string, notated $1$, and of strings of the form $x_1 x_2 ... x_n$,
where $x_i$ is an atom on $V$. It is assumed that such a string is
\conceptdef{reduced}, that is, never contains two consecutive atoms
which are inverse of each other: no substring $v v\i$ or $v\i v$ is
allowed to appear in a reduced string.

When $\alpha$ and $\beta$ are two reduced strings, their concatenation
$\alpha \beta$ can be reduced by eliminating all substrings of the
form $v v\i$ or $v\i v$. It can be proven that the reduced string
$\gamma$ obtained in this way is independent of the order of such
eliminations.  In this way, a product on $\freegroup{V}$ is defined,
and it is easily shown that $\freegroup{V}$ becomes a
(non-commutative) group, called the \conceptdef{free group} over $V$
\cite{Rotman:1994}.

\subsection{Group computation} 
We will say that an ordered pair
$\gcs{} = (V,R)$ is a \conceptdef{group computation structure} if:
\begin{enumerate}
\item $V$ is a set, called the \conceptdef{vocabulary}, or the set of
  \conceptdef{generators}
\item $R$ is a subset of $\fg$, called the \conceptdef{lexicon}, or
  the set of \conceptdef{relators}.\footnote{For readers familiar with
    group theory, this terminology will evoke the classical notion of
    group \conceptdef{presentation} through generators and relators.
    The main difference with our definition is that, in the classical
    case, the set of relators is taken to be symmetrical, that is, to
    contain $r\i$ if it contains $r$. When this additional assumption
    is made, $\nm{R}$ becomes equal to $\ng{R}$.}
\end{enumerate}

\sloppy
The submonoid closure $\nm{R}$ of $R$ in $\fg$ is called
the \conceptdef{result \linebreak monoid} of the group computation structure
$\gcs{}$. The elements of $\nm{R}$ will be called \conceptdef{computation
results}, or simply \conceptdef{results}.
\fussy

If $r$ is a relator, and if $\alpha$ is an arbitrary
element of $\fg$, then $\alpha r \alpha\i$ will be called
a \conceptdef{quasi-relator} of the group computation structure. It is
easily seen that the set $R_N$ of quasi-relators is equal
to the normal subset closure of $R$ in $\fg$, and that
$\nm{R_N}$ is equal to $\nm{R}$.

A \conceptdef{computation} relative to $\gcs{}$ is a finite sequence $c = 
(r_1,
\ldots, r_n)$ of quasi-relators. The product $r_1 \cdots r_n$ in $\fg$
is evidently a result, and is called the \conceptdef{result of the 
computation}
$c$. It can be shown that the result monoid is
entirely covered in this way: each result is the result of
some computation. A computation can thus be seen as a ``witness'', or 
as
a ``proof'', of the fact that a given element of $\fg$ is a result of
the computation structure.\footnote{The analogy with the view in
  constructive logics is clear. There what we call a result is called
  a {\em formula} or a {\em type}, and what we call a computation is called 
a {\em proof}.}

For specific computation tasks, one focusses on results of a certain
sort, for instance results which express a relationship of input-output,
where input and output are assumed to belong to certain object types.
For example, in computational linguistics, one is often interested in
results which express a relationship between a fixed semantic input and a
possible textual output (generation mode), or conversely in results
which express a relationship between a fixed textual input and a
possible semantic output (parsing mode).

If $\gcs{} = (V,R)$ is a group computation structure, and if
$A$ is a given subset of $\fg$, then we will call the pair
$\gcsa{} = (\gcs{}, A)$ a \conceptdef{group computation structure with
acceptors}. We will say that $A$ is the set of acceptors,
or the \conceptdef{public interface}, of $\gcsa{}$. A result of $\gcs{}$
which belongs to the public interface will be called a
\conceptdef{public result} of $\gcsa{}$.

\section{Relating the geometric and the algebraic views: the
  fundamental theorem of combinatorial group theory}

The two views of computation provided on the one hand by diagrams and
on the other hand by group computation structures are in fact
equivalent to each other. This equivalence is the consequence of the
``the fundamental theorem of combinatorial group theory''
\cite{Rotman:1994}.

\subsection{Cyclically reduced words}

{\bf Definition}. A word $w$ on \atoms\ is said to be \conceptdef{cyclically
  reduced} iff every cyclic permutation of it is reduced. 

It is easy to see that:
\begin{itemize}
\item A reduced word is cyclically reduced iff it is not of the form
  $a w' a\i$ with $a$ an atom (positive or negative);
\item If a word is cyclically reduced, then all its cyclic
  permutations are cyclically reduced;
\item For any word $w$, there is a conjugate of $w$ which is
  cyclically reduced;
\item Two conjugates of a word $w$ which are cyclically reduced are
  cyclic permutations of each other.
\end{itemize}

It is often convenient to picture the set of all cyclic permutations
of a cyclically reduced word as a circular diagram of labelled oriented
edges such that no adjacent edges cancel each other. For any word $w$, 
such a diagram provides a canonical representation of the cyclically
reduced conjugates of $w$.

\subsection{Relator cells} \label{relator-cells}

Consider a group computation structure $\gcs{} = (V,R)$. Without loss of
generality, it can be assumed that the relators in $R$ are cyclically
reduced, because the result monoid is invariant when one considers a
new set of relators consisting of conjugates of the original ones. From 
now on, unless stated otherwise, this assumption will be made 
for all relators considered. 

Take any such cyclically reduced relator $r = x_1^{e_1} \ldots
x_n^{e_n}$, where $x_i \in V$ and $e_i = \pm 1$, and construct
a labelled cell in the following way: take a circle and divide it in
$n$ arcs; label the clockwise-$i$th arc $x_i$ and orient it clockwise
if $e_i =1$, anti-clockwise otherwise. The labelled cell thus obtained
is call the \conceptdef{relator cell} associated with $r$.  

Rather than presenting the \gcs{} through a set of relator words as we
have done before, it is now possible to present it through a set of
relator cells; if one gives such a set, a standard presentation of the
\gcs{} can be derived by taking an arbitrary origin on each cell and
``reading'' the relator word clockwise from this origin; the origin
chosen does not matter: any other origin leads to a conjugate
relator, and this does not affect the notion of result. 

\subsection{Fundamental theorem of combinatorial group theory}

We are now able to state what J. Rotman calls ``the fundamental
theorem of combinatorial group theory'' \cite{Rotman:1994}. We give
the theorem in a slightly extended form, adapted to the case of a
\gcs{}, that is, using normal {\em sub-monoid} closure rather than
normal {\em subgroup} closure; the subgroup case follows immediately
by taking a set or relators containing $r\i$ along with $r$.

\begin{theorem}\label{fundamental theorem of CGT} Let $\gcs{} = (V,R)$ be a 
group computation structure such that
all relators $r\in R$ are cyclically reduced. If $w$ is a cyclically
reduced word in \fg, then $w \in \nm{R}$ if and only if there exists
a reduced diagram having boundary word $w$ and whose regions are
relator cells associated with the elements of $R$.
\end{theorem}

\begin{figure}[t]
  \begin{center}
    \epsfig{file=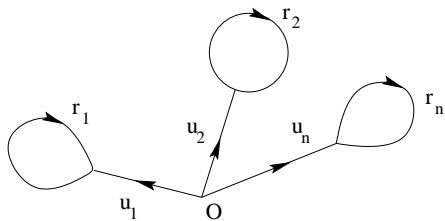,scale=0.8}
    \caption{Star diagram.}
    \label{fig:star-diagram}
  \end{center}
\end{figure}

The proof is not provided; it can easily be recovered from the
property demonstrated in \cite{LyndonSchuppe:1977} (chapter 5, Section
1). The proof involves the following remark. If one considers a
product
$${u_1}r_1{u_1}\i\ldots {u_n}r_n{u_n}\i$$ 
with $r\i\in R$ and $u_i$ arbitrary elements of \fg, this product can
be read as the boundary word of the ``star'' diagram represented in
\myfigref{fig:star-diagram}, starting at $O$ and progressing clockwise. 

This star diagram is in general not in reduced form, but it can be
reduced by a stepwise process of ``stitching together'' edges which do
not respect the definition of a reduced diagram.

{\bf Example}. Let's consider a \gcs{} with vocabulary $V = \set{a,b,c}$ and
set of (cyclically reduced) relators  
$$R = \set{c\i{}c\i{}a\i{}c\i{}, acb\i, baa}$$.

\begin{figure}[htp]
  \begin{center}
    \epsfig{file=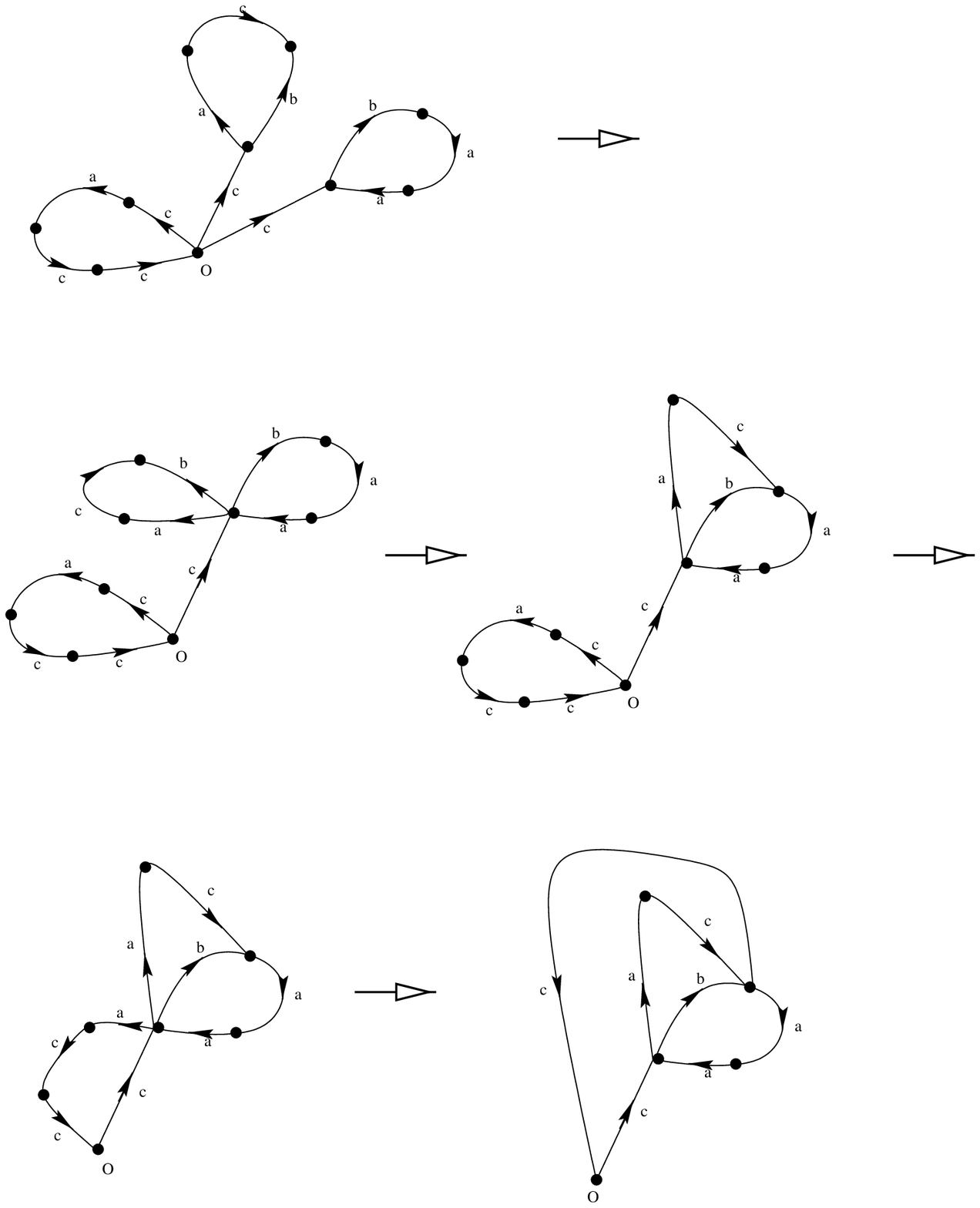,scale=0.7}
    \caption{Transformation of a diagram into reduced form (adapted
      from {\protect \cite{LyndonSchuppe:1977}}).}
    \label{fig:diagram-reduction}
  \end{center}
\end{figure}

The cyclically reduced word $c\i a a c\i$ is an element of \nm{R}, for
it can be obtained by forming the product  
$$c\i{}c\i{}a\i{}c\i \;\;\; c acb\i c\i \;\;\; c baa c\i.$$  
If we form the star diagram for this product, we obtain the first
diagram shown in \myfigref{fig:diagram-reduction}. 

This diagram is not reduced, for instance the two straight edges
labelled $c$ are offending the reduction condition. If one
``stitches'' these two edges together, one obtains the second diagram
in the figure. This stitching corresponds to a one-step reduction of
the boundary word of the first diagram, $c\i{}c\i{}a\i{}c\i{} \; c 
acb\i{} c\i
\; c baa c\i$ into the boundary word of the second 
\linebreak 
$c\i{}c\i{}a\i{}c\i{} \; c acb\i{} baa c\i$. By continuing in this way, 
one obtains the fifth
diagram of the figure, which is reduced, and whose boundary is the
desired result $c\i a a c\i$. 

\section{Applications to grammar}

\subsection{A simple grammar}

\begin{figure}[h]
  \begin{center}
    \epsfig{file=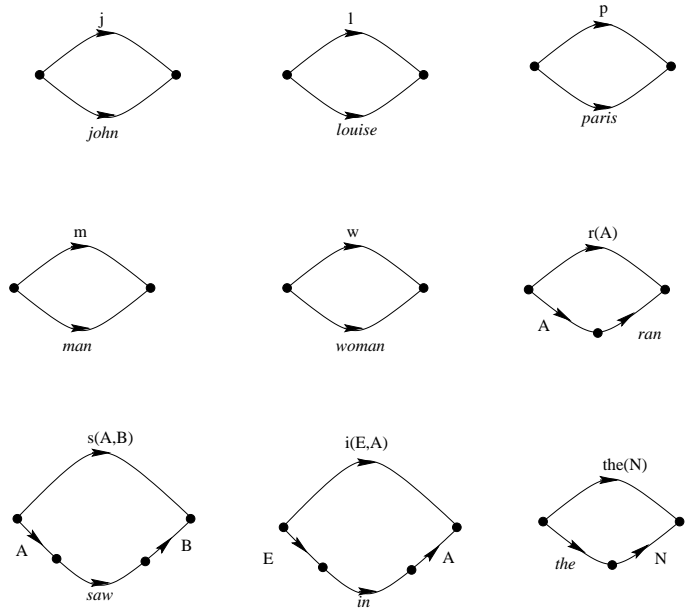,width=9cm}
    \caption{\ggone{} given in geometric terms: cells schemes.}
    \label{fig:linguistic-cells}
  \end{center}
\end{figure}

We will now show how the formal concepts introduced above can be
applied to the problems of grammatical description and computation. We
start by introducing a simple grammar \ggone{} for a fragment of
English. In Figure \ref{fig:g-grammar}, this grammar is presented
algebraically in terms of relators, and in Figure
\ref{fig:linguistic-cells} the same grammar is presented geometrically
in terms of relator cells.

Formally, \ggone{} is a group computation structure with acceptors
over a vocabulary $V = V_{log} \union V_{phon}$ consisting of a set of
logical forms $V_{log}$ and a disjoint set of phonological elements
(in the example, words) $V_{phon}$. Thus
{\tt \w{john}}, {\tt \w{saw}} are  phonological elements,  {\tt j},\, {\tt
  s(j,l)} are logical forms.

The grammar lexicon, or set of relators, $R$ is given as a list of
``lexical schemes''. Thus, in Figure \ref{fig:g-grammar}, each line is a
lexical scheme and represents a set of relators in $\fg$. The first
line is a ground scheme, which corresponds to the single relator {\tt
  j \w{john}\i}, and so are the next four lines. The sixth line is a
non-ground scheme, which corresponds to an infinite set of relators,
obtained by instanciating the {\em term meta-variable} {\tt A}
(notated in uppercase) to a logical form. So are the remaining lines.

The vocabulary and the set of relators that we have just specified
define a group computation structure $\gcs{} = (V,R)$. Let's now
describe a set of acceptors $A$ for this computation structure. We
take $A$ to be the set of elements of $\fg$ which are products of the
following form:
$$S \: {W_n}\i {W_{n-1}}\i \ldots {W_1}\i$$
where $S$ is a logical
form ($S$ stands for ``semantics''), and where each $W_i$ is a
phonological element ($W$ stands for ``word''). The expression above
is a way of encoding the ordered pair consisting of the logical form
$S$ and the phonological string $W_1 W_2 \ldots W_n$ (that is, the
inverse of the product ${W_n}\i {W_{n-1}}\i \ldots {W_1}\i$).

\newpage

\vspace*{10mm}

\begin{figure}[h]
  \begin{center}
\begin{alltt}
                j \w{john}\i
                l \w{louise}\i
                p \w{paris}\i
                m \w{man}\i
                w \w{woman}\i
                r(A) \w{ran}\i A\i 
                s(A,B) B\i \w{saw}\i A\i 
                i(E,A) A\i \w{in}\i E\i
                t(N) N\i \w{the}\i
\end{alltt}    
\caption{\ggone{} given in algebraic terms: relator schemes}
\label{fig:g-grammar}
  \end{center}
\end{figure}

\vspace*{5mm}

\clearpage

\subsection{Computation}

We now show how to compute a proof of the fact that {\tt i(s(j,l),p)} \linebreak
$\w{paris}\i \w{in}\i \w{louise}\i \w{saw}\i \w{john}\i$ is a public
result for \ggone, or, in other words, that the logical form
$i(s(j,l),p)$ and the phonological form  \w{john
saw louise in paris} are in correspondance relative to \ggone.

We start by an informal geometric computation of this result and
follow by an algebraic computation of it. The geometric computation if
the more intuitive of the two, the algebraic one the one for which we
have the more precise formal definition and which more directly
displays group-theoretical characteristics.

\subsubsection{Geometric computation}

\begin{figure}[th]
  \begin{center}
    \epsfig{file=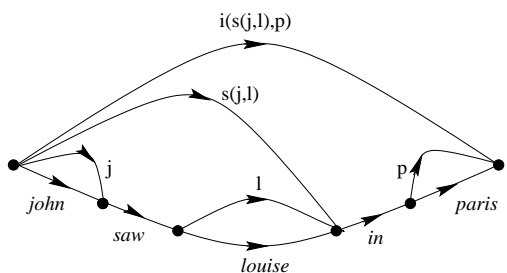,scale=0.85}
    \caption{A diagram establishing the relationship between a logical
      form and a phonological string.}
    \label{fig:parsing-generation-diagram}
  \end{center}
\end{figure}

Consider the diagram of Figure \ref{fig:parsing-generation-diagram}.
The cells of this diagram are instances of the cell schemes of Figure
\ref{fig:linguistic-cells}. From theorem \ref{fundamental theorem of
  CGT}, this means that {\tt i(s(j,l),p)} $\w{paris}\i \w{in}\i
\w{louise}\i \w{saw}\i \w{john}\i$ is a public result of \ggone.

The diagram in Figure \ref{fig:parsing-generation-diagram} can be
obtained by several computations, in the sense of Section
\ref{Diagrams as computational devices}. One such computation,
corresponding to a generation mode, starts from the top edge ({\tt
  i(s(j,l),p)}) and progressively adds cells in a top-down way. A
second computation, corresponding to a parsing mode, starts by
building a diagram consisting of the sequence of the five phonological
edges at the bottom and progressively adds cells in a bottom-up way.
Still another geometric computation, establishing the connection with
the algebraic computation, will be shown below.

\subsubsection{Algebraic computation}

\noindent Consider the following relators, instanciations of relator
schemes of Figure \ref{fig:g-grammar}:
\begin{alltt}
r\sub1 = i(s(j,l),p) p\i \w{in}\i s(j,l)\i
r\sub2 = s(j,l) l\i \w{saw}\i j\i
r\sub3 = j \w{john}\i
r\sub4 = l \w{louise}\i
r\sub5 = p \w{paris}\i
\end{alltt}
and the quasi-relators:
\begin{alltt}
r\sub1' = r\sub1
r\sub2' = r\sub2
r\sub3' = r\sub3
r\sub4' = (\w{john} \w{saw}) r\sub4 (\w{john} \w{saw})\i
r\sub5' = (\w{john} \w{saw} \w{louise} \w{in}) r\sub5 (\w{john} \w{saw} \w{louise} \w{in})\i
\end{alltt}
Then we have:
\begin{alltt}
r\sub1' r\sub2' r\sub3' r\sub4' r\sub5' =
  i(s(j,l),p) \w{paris}\i \w{in}\i \w{louise}\i \w{saw}\i \w{john}\i
\end{alltt}
and therefore {\tt i(s(j,l),p) \w{paris}\i \w{in}\i \w{louise}\i
  \w{saw}\i \w{john}\i} is the result of a computation {\tt
  (r\sub1',r\sub2',r\sub3',r\sub4',r\sub5')}, as announced.

\begin{figure}[th]
  \begin{center}
    \epsfig{file=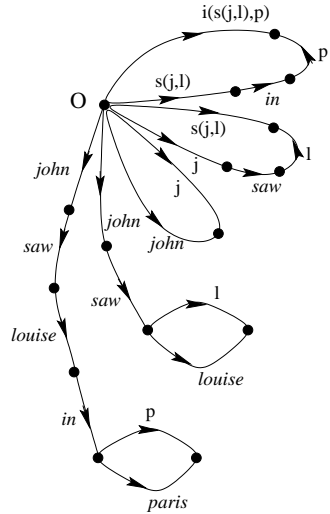,scale=0.8}
    \caption{An unreduced version of the diagram of Figure
      \protect\ref{fig:parsing-generation-diagram}, showing
      explicitly the role of conjugates.}
    \label{fig:john-saw-star}
  \end{center}
\end{figure}

What is the relationship of this computation with the geometric view
of Figure \ref{fig:parsing-generation-diagram}? The answer is given by
the geometric computation indicated by the diagram in Figure
\ref{fig:john-saw-star}. If we read the boundary of this diagram
clockwise starting from O, we obtain the unreduced expression {r\sub1'
  r\sub2' r\sub3' r\sub4' r\sub5'}. But if we perform further
geometric reduction steps on the diagram of Figure
\ref{fig:john-saw-star}, we obtain the diagram of Figure
\ref{fig:parsing-generation-diagram}.

\section{Multi-relators and non-local dependencies}

\subsection{Multi-relators}

The linguistic examples that we have presented up to now have been
rather simple. We will now introduce an extension of group computation
structure which gives rise to forms of non-local grammar dependencies.

Let $\mr = \multirelator{w_1 ; \ldots ; w_n}$ be a finite multiset
(unordered list) of words in $(\atoms)^*$. We will call such an
expression a \conceptdef{multi-relator} over $V$. We will say that a
word $w$ is a \conceptdef{multi-conjugate} of $\mr$ iff $w$ can be
expressed as a product ${\alpha_1} w_1 {\alpha_1}\i \ldots {\alpha_n}
w_n {\alpha_n}\i$, where $\alpha_1, \ldots, \alpha_n$ are elements of
$\fg$. Let $\MC(\mr)$ be the set of multi-conjugates of $\mr$.

{\em Remark:} This notion is well-defined, for any ordering of the multiset
$\mr$ leads to the same set of multi-conjugates (by a simple property
of conjugacy).  It is also easy to check that replacing any $w_i$ by a 
cyclic permutation of $w_i$ does not change $\MC(\mr)$, and
furthermore, that $\MC(\mr)$ is a normal subset of $\fg$.

If we are given a (finite or infinite) collection ${MR}$ of
multi-relators, we can consider the group computation structure
obtained by taking as set $R$ of relators the set:
$$R = \bigcup_{{\mr}\in {MR}} \multiconjugates{\mr};$$

We will call the GCS thus obtained the \conceptdef{group computation 
structure with multi-relators} $ \mbox{GCS{-}MR} = (V,{MR})$.

When presenting a group computation structure with multi-relators, we
will sometimes use the notation:
$$w_1 ; w_2 ; \ldots ; w_n$$
for presenting a multi-relator. A
multi-relator consisting of only one relator $w_1$ will be written
simply $w_1$; specifying such a multi-relator has exactly the same
effect on the computation structure as specifying the simple relator
$w_i$, because $\MC(\multirelator{w_1})$ is normal in \fg.

\subsection{Multi-relators, multi-cells, and diagrams}

We will now state a theorem which is an extension of the the
fundamental theorem of combinatorial group theory for the case of
GCS's with multi-relators. We first need the notion of multi-cell
associated with a multi-relator.

We first remark that we can always assume that a multi-relator
$\multirelator{w_1 ; w_2 ; \ldots ; w_n}$ is such that each $w_i$ is
cyclically reduced, because taking the cyclically reduced conjugate of
$w_i$ does not change the notion of multi-conjugate. We will assume
this is always the case in the sequel.

If $\multirelator{w_1 ; w_2 ; \ldots ; w_n}$ is a multi-relator, and
if $\Gamma_i$ is the cell associated with $w_i$ in the manner of
\ref{relator-cells}, then we will call the multiset of cells \linebreak
$\Theta =
\multiset{\Gamma_1, \ldots, \Gamma_n}$ the \conceptdef{multi-cell}
associated with this multi-relator.

Consider a finite multiset of multi-relators and take the multi-set 
obtained by forming the multiset union $\Omega = \biguplus_{k} \Theta_k$ of 
the multi-cells associated with these multi-relators. A diagram whose 
cells are exactly (that is, taking account of the cell counts) those of 
the multiset $\Omega$ will be said to be a \conceptdef{diagram relative 
to} the GCS with multi-relators under consideration.

One can easily prove the following extension of theorem \ref{fundamental theorem of CGT}.
\begin{theorem}\label{extended fundamental theorem of CGT}. 
Let $\mbox{GCS{-}MR} = (V,{MR})$ be a group computation structure with 
multi-relators. If $w$ is a cyclically reduced word in \fg, then $w$ is in the 
result monoid of $\mbox{GCS{-}MR}$ iff there exists a reduced diagram 
relative to 
$\mbox{GCS{-}MR}$ having boundary $w$.
\end{theorem}

\subsection{Linguistic examples}

Let's consider the extension \ggtwo{} of \ggone{} presented in Figure
\ref{fig:g-grammar with multi-relators}. The first nine entries are
the same as before, but three new multi-relator schemes have been
added to the end. They correspond to definitions of the quantifiers
``every'' and ``some'' and to the definition of the relative pronoun
``that''. 

The notation {\tt P[\x]} is employed to express the fact that a
logical form containing an {\em argument identifier}\  {\tt \x} is equal to 
the
application of the abstraction {\tt P} to {\tt \x}. The {\em identifier 
meta-variable}
{\tt X} in {\tt P[X]} ranges over such identifiers ({\tt \x}, {\tt
  \y}, {\tt \z}, ...), which are notated in lower-case italics (and are 
always ground).  

The meta-variable {\tt P} ranges over logical form abstractions
missing one argument (for instance {\tt\la\,\z.s(j,\z)}).  When
matching meta-variables in logical forms, use of higher-order
unification will be allowed. For instance, one can match {\tt P[X]} to
{\tt s(j,\x)} by taking ${\tt P} = {\tt \la\,\z.s(j,\z)}$ and ${\tt X}
= {\tt \x}$.

\begin{figure}[t]
  \begin{center}
\begin{alltt}
                j \w{john}\i
                l \w{louise}\i
                p \w{paris}\i
                m \w{man}\i
                w \w{woman}\i
                r(A) \w{ran}\i A\i 
                s(A,B) B\i \w{saw}\i A\i 
                i(E,A) A\i \w{in}\i E\i 
                t(N) N\i \w{the}\i
                ev(N,X,P[X]) P[X]\i ; X N\i \w{every}\i
                sm(N,X,P[X]) P[X]\i ; X N\i \w{some}\i
                tt(N,X,P[X]) P[X]\i \w{that}\i N\i ; X
\end{alltt}    
    \caption{\ggtwo{} given in algebraic terms.}
    \label{fig:g-grammar with multi-relators}
  \end{center}
\end{figure}

\begin{figure}[th]
  \begin{center}
    \epsfig{file=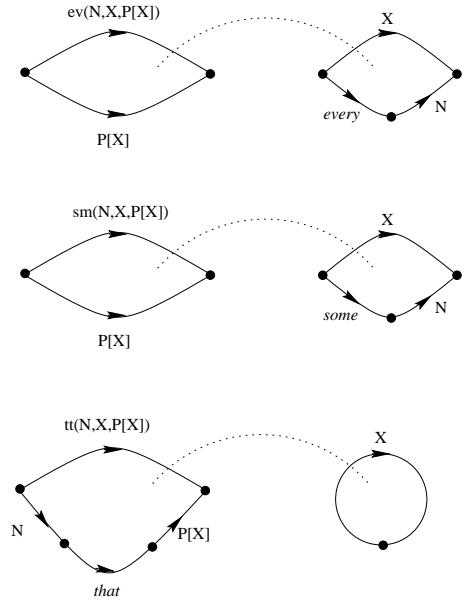,width=6cm}
    \caption{\ggtwo{} given in geometric terms (omitting the cells
      already described in Figure \ref{fig:linguistic-cells}).}
    \label{fig:multi-cells}
  \end{center}
\end{figure}

The geometric presentation of \ggtwo{} is given in Figure
\ref{fig:multi-cells}, for the three new multi-relators (we have
omitted reproducing the cells of Figure \ref{fig:linguistic-cells},
which also belong to the specification). The dotted lines indicate
which cells belong to a given multi-cell. The cells belonging to a
multi-cell always work in solidarity: they have to appear
together in a diagram or not at all.

\subsection{Linguistic computation with multi-relators}

\begin{figure}[H]
  \begin{center}
    \epsfig{file=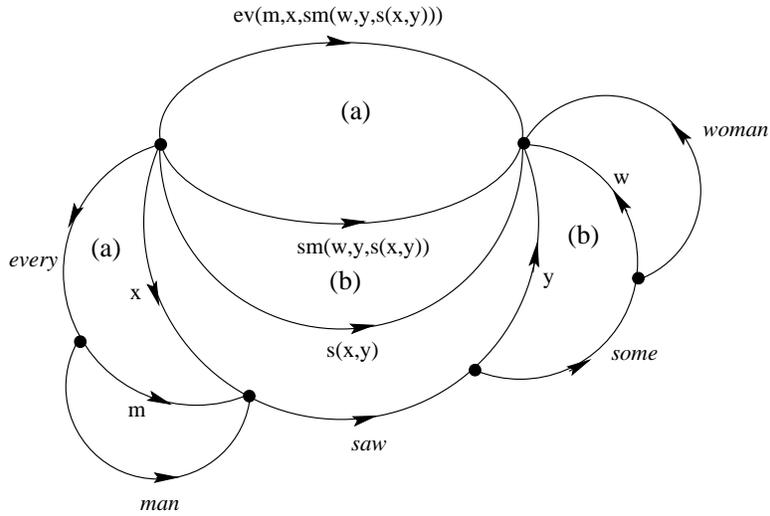,scale=.95}
    \caption{A diagram using multi-cells establishing a correspondence
      between the sentence \w{every}\ \w{man}\ \w{saw}\ \w{some}\ 
      \w{woman} and the logical form {\tt ev(m,\x,sm(w,\y,s(\x,\y)))}.}
    \label{fig:every-some}
  \end{center}
\end{figure}

\begin{figure}[H]
  \begin{center}
    \epsfig{file=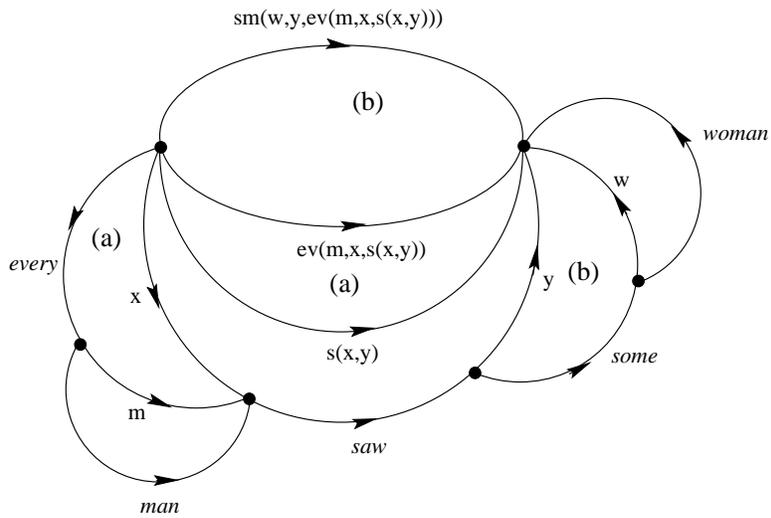,scale=.95}
        <\caption{A diagram establishing the correspondence between the
          same sentence and the differently scoped logical form {\tt
            sm(w,\y,ev(m,\x,s(\x,\y)))}.}
    \label{fig:some-every}
  \end{center}
\end{figure}

\begin{figure}[h]
  \begin{center}
    \epsfig{file=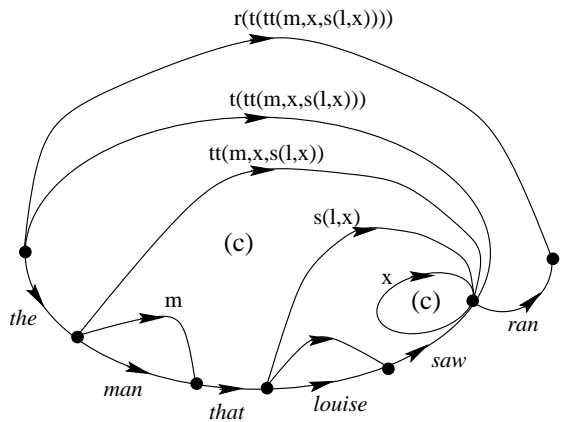,scale=.95}
    \caption{A diagram using multi-cells establishing a correspondence
      between the sentence \w{the man that louise saw ran} and the
      logical form {\tt r(t(tt(m,x,s(l,x))))}.}
    \label{fig:man-that}
  \end{center}
\end{figure}

The figures \ref{fig:every-some}, \ref{fig:some-every}, and
\ref{fig:man-that} illustrate some computations with
\ggtwo.\footnote{For a more detailed discussion, and for the algebraic 
  counterparts of these computations, see \cite{Dymetman:1998c}.}

\sloppy

The first diagram consists of seven cells: three ``mono''-cells, and
two ``bi''-cells (a) and (b), corresponding respectively to the
entries for ``every'' and for ``some'' in the grammar. The boundary of
the diagram can be read {\tt ev(m,\x,sm(w,\y,s(\x,\y)))} \w{woman\i
  some\i saw\i man\i every\i}. Because of theorem \ref{extended
  fundamental theorem of CGT}, this proves the correspondence relative
to \ggtwo{} of the logical form {\tt ev(m,\x,sm(w,\y,s(\x,\y)))} with
the sentence \w{every}\ \w{man}\ \w{saw}\ \w{some}\ \w{woman}.

\fussy

The second diagram is similar to the first, but with a different
layout between the multi-cells for ``every'' and ``some''. It
establishes a different scoping for the quantifiers of the same
sentence.

The third diagram establishes the correspondence between the sentence
\w{the man that louise saw ran} and the logical form {\tt
  r(t(tt(m,x,s(l,x))))}.

\newpage
\clearpage

\section{Final remarks}

\subsection{Normal submonoids versus subgroups}

In Section \ref{Groups, monoids, normal subsets}, we stated without
justification that, for the purpose of presenting group computation
structures, we would use normal submonoids of a free group rather than
normal subgroups, which are much more central in algebra. We would
like now to give some support to this claim.

Suppose that we were to define computation results as being elements
of the normal subgroup closure $\ng{R}$ of a set of relators, rather
than of the normal submonoid closure $\nm{R}$. Then $\ng{R}$ would
partition the elements of \fg{} into equivalence classes: two elements
$x,y\in\fg$ being equivalent iff $x y\i \in \ng{R}$. For all purposes
two such elements would become undistinguishable. For linguistic
formalization, where we need to model the relationship between a
logical form and a phonological form, this approach would neutralize
important differences: if we had, say, the results $\mit sem_1
string_1\i$, $\mit sem_2 string_1\i$, and $\mit sem_2 string_2\i$ ---
meaning that $\mit sem_1$ is associated with $\mit string_1$, and
$\mit sem_2$ is associated with both $\mit string_1$ and $\mit
string_2$, then $\mit sem_1, sem_2, string_1, string_2$ would all be
in the same equivalence class, meaning in particular that $\mit sem_1$
would be associated
with $\mit string_2$, counter-intuitively.%
\footnote{Of course, if it is deemed useful to identify two
  expressions completely, such as in case of synonymy between two
  logical forms $\mit sem_a$ and $\mit sem_b$, it would be possible to
  specify both relators $sem_a sem_b\i$ and $sem_b sem_a\i$ in the
  group computation structure. One might then perhaps even think of
  {\em quotienting} the group computation structure by the equivalence
  relation thus created.}  This problem does not occur with the choice
of $\nm{R}$: in this case the relation between $x,y$ defined by $x y\i
\in \nm{R}$ is only a preorder, not an equivalence relation
(see \cite{Dymetman:1998c}).

\subsection{Group computation versus rewriting}

We started this paper in Section \ref{Grammar and geometry} by taking a
geometric viewpoint on conventional rewriting systems such as
context-free grammars or type-0 grammars. Before ending, we would like
to consider the following question: what, if any, is the difference
between group computation and these rewriting systems? A detailed
discussion of this question is outside the scope of this paper (see
\cite{Dymetman:1998c}), and only some brief indications will be given
here.

Let's consider a type-0 rewriting system and a type-0 chart $\mit
CH$ relative to this system. We have seen in Section \ref{Grammar
  and geometry} that such a chart can be obtained by ``pasting''
together cells corresponding to the rules of the rewriting system.
That is, the chart $\mit CH$ can be seen as a diagram relative to
the \gcs{} defined by this collection of cells. But such a diagram has
an interesting property among all diagrams that could be produced
relative to the \gcs{}: the cells of $\mit CH$ are partially ordered
by the relation which considers a cell $c_1$ of a diagram to
immediately precede a cell $c_2$ of the same diagram iff there is a
common edge between $c_1$ and $c_2$ which is negatively oriented
relative to $c_1$ and positively oriented relative to $c_2$. For a
standard presentation of the chart $\mit CH$, this partial order is
just the usual ``top-down'' order between the
cells.\footnote{Remark: the several
  {\em total orders} which are compatible with the partial order of
  precedence in a chart are in one-to-one correspondence with the
  {\em derivations} associated with the chart.}

The fact that the cells of the diagram $\mit CH$ are
partially-ordered by the precedence relation just defined is a {\em
  global} property of the diagram which means that the precedence
relation does not create cycles among cells. It is a property which is
foreign to group computation, where cells are assembled relative to a
purely {\em local} criterion.\footnote{One could say that group
  computation ignores the directionality of time, contrarily to a
  rewriting system.}
 This means that, in general, the
translation of a type-0 system into a \gcs{} produces diagrams that
cannot be interpreted as derivations of the original type-0 system.
However, in the case where the relator cells resulting from the
translation of the rewriting system can be {\em statically}
partially-ordered by the precedence relation where $c < c'$ iff $c$ has a
negatively oriented edge labelled $l$ and $c$ has a positively
oriented edge with the same label $l$, {\em then} no cycles can ever
appear in diagrams of the \gcs{}, and then the two notions of
computation become identical.

The static condition just described is very restrictive for type-0
systems {\em stricto sensu}, that is, having a finite vocabulary of
non-terminals, because it precludes recursivity. It is, however,
much more interesting in the case of rewriting systems defined by
rule-schemes over terms, such as DCGs \cite{PeWa80}, or their type-0
counterparts. An example of such a situation is provided by the cells
of Figure \ref{fig:linguistic-cells}, which can be seen to correspond
to an ``extended'' type-0 system of this kind, and where the
instanciated relator cells can be partially ordered by the precedence
relation just described (see \cite{Dymetman:1998c}).

\subsection{Group morphisms and computational complexity}

One important consequence of working within a group-theoretical
framework is that group theory provides powerful tools for studying
invariant properties of its objects. 

Let us give an illustration of this fact on the basis of our example
grammar \ggtwo{} of Figure \ref{fig:g-grammar with multi-relators}.

Let's first define the {\em semantic size} $\ss$ of a logical form
term as the number of nodes in this term which are different from
argument identifiers such as {\tt \x}, {\tt \y}, {\tt \z}... (thus,
$\ss({\tt ev(m,\x,sm(w,\y,s(\x,\y)))}) = 5$), and the semantic size of
a phonological word as 0. Let's also define the {\em phonological
  size} $\ps$ of a phonological word as 1, and the phonological size
of a logical form as 0.

The functions $\ss$ and $\ps$ can be extended to morphisms from \fg{}
to $\Z{}$ in the standard way. Let's then consider the morphism $h$ from
\fg{} to $(\Z \times \Z, +)$ defined by $h(w) = (\ss(w),\ps(w))$.

If one looks at the multi-relator schemes of Figure \ref{fig:g-grammar with
  multi-relators}, it can be checked that any grounded multi-conjugate 
instance $w$ of each multi-relator scheme is such that $h(w) =
(1,-1)$. For example, taking the multi-relator scheme {\tt
  ev(N,X,P[X]) P[X]\i ; X N\i \w{every}\i} we see that any
multi-conjugate instance $w$ of this scheme is such that 
$$h(w) =
h({\tt ev(N,X,P[X])}) + h({\tt P[X]\i}) + h({\tt X}) + h({\tt N\i}) +
h(\w{every}\i),$$
because taking conjugates does not change the value of a morphism like 
$h$ which takes its values in a commutative group. Thus we have:
$$h(w) = (1 + \ss({\tt N}) + 0 + \ss({\tt P[X]}) - \ss({\tt P[X]}) + 0
- \ss({\tt N}), -1) = (1,-1),$$
and so on for the other relator
schemes.

Consider now a computation of a result $w$ involving $n$
multi-relators (in geometric terms, a diagram with boundary $w$
involving $n$ multi-cells). The value of $h(w)$ is then $(1,-1) +
\ldots + (1,-1)$ taken $n$ times, that is, $h(w) = (n,-n)$. This has
the following consequences (stated informally here) for the complexity
of parsing and generation:\footnote{In the terminology of
  \cite{DymKluwer94}, the properties 1,2 and 3 mean that the grammar
  is {\em inherently reversible}.}
\begin{enumerate}
\item If a phonological string and a logical form are in
  correspondence relative to the grammar, then it is possible to bound
  the phonological size of the string as a function of the semantic
  size of the logical form (in fact they are equal).
\item If we have a string of phonological size $n$ to parse, then any
  computation will involve at most (and in fact exactly) $n$
  multi-cells; this implies that parsing is decidable and of bounded complexity.
\item If we have a logical form of semantic size $n$ to generate, then
  any computation will involve at most (and in fact exactly) $n$
  multi-cells; this implies that generation is decidable and of
  bounded complexity.
\end{enumerate}

Although the grammar \ggtwo{} is rather remarkable in admitting a
morphism such as $h$ which has the same value on all the
multi-relators, what is really needed for the complexity properties to
hold is a less demanding requirement (see \cite{Dymetman:1998c}) of
finding a morphism which realize some reasonable ``exchange'' of
phonological material for semantic material on each multi-relator.


\end{document}